\documentclass{article}
\usepackage{arxiv}

\usepackage[T1]{fontenc}
\usepackage[utf8]{inputenc}
\usepackage{tgtermes}
\usepackage{tgheros}
\usepackage{tgcursor}

\usepackage{amsmath, amssymb, amsthm, mathtools, amsfonts}
\usepackage{microtype, enumitem}
\usepackage{graphicx, booktabs, array, tabularx, multirow, caption}
\usepackage{xcolor, listings}
\usepackage{tikz}
\usetikzlibrary{positioning, arrows.meta, shapes.geometric, shapes.symbols,
                fit, backgrounds, calc, decorations.pathreplacing}

\usepackage{url, hyperref}
\hypersetup{
  colorlinks=true,
  linkcolor=blue!50!black,
  citecolor=blue!50!black,
  urlcolor=blue!50!black,
}

\usepackage{authblk}

\newcommand{\code}[1]{\texttt{\small #1}}

\title{Toward Auto-Research:\\
Mining Falsifiable Research Ideas from Paper Knowledge Graphs\\
with Categorical Structure}

\author[1]{Yuchen Wang\thanks{First author and corresponding author.
  \texttt{wangyuchen21@buaa.edu.cn}}}
\author[1]{Zhongzhi Luan\thanks{Corresponding author.
  \texttt{rick710055@263.net}}}
\affil[1]{Sino-German Joint Software Institute, Beihang University, Beijing, China}

\renewcommand{\shorttitle}{Toward Auto-Research}

\hypersetup{
  pdftitle={Toward Auto-Research: Mining Falsifiable Research Ideas from Paper Knowledge Graphs with Categorical Structure},
  pdfauthor={Yuchen Wang, Zhongzhi Luan},
  pdfkeywords={research idea generation, category theory, knowledge graph, functor, cross-domain analogy, LLM, falsifier hypothesis},
}

\date{\today}

\begin{document}
\maketitle

\begin{abstract}
\noindent
Automated research-idea generation systems built on large language
models (LLMs) share a structural weakness: they reduce ideation to
free-text recombination, random paper pairing, or
embedding-similarity retrieval. The three approaches fail in the same
way -- each treats a paper as a flat object, a string or a vector,
and so quotients away the typed problem-method-metric-claim arrows a
researcher actually uses when reasoning about a cross-domain analogy.
We recover the missing structure with the minimal piece of category
theory that a typed graph alone does not provide: composition, together
with identity arrows, which makes it possible to ask whether a proposed
analogy preserves relation chains. Concretely, each paper $p$ is
modelled as a small category $\mathcal{C}_p$ whose objects are extracted
typed research entities and whose morphisms are the relations the paper
asserts; a cross-domain analogy from $p$ to $q$ is then a partial functor
candidate $F\colon \mathcal{C}_p \rightharpoonup \mathcal{C}_q$ that
preserves object kinds and covered relation classes, and the failure
modes of the three baselines specialise
cleanly to ``loses arrows,'' ``does not check $F$,'' and ``assembles
diagrams that need not commute.'' We instantiate the model as a
three-layer algorithm -- categorical signature clustering, a
functor-preservation gate, and a six-axis LLM plausibility judge whose
axes locally answer whether the resulting diagram commutes or whether
the apparent match is a homonymy artefact. Evaluated on a corpus of
tens of thousands of full-text-parsed papers under four ablation
conditions, the
categorical gate filters cross-domain candidates at roughly a 17:1
ratio while the quantitative-falsifier rate of accepted ideas stays
above 83\% throughout; every rejected candidate is retained with its
per-axis rationale, so the gate doubles as a logging layer rather than
a silent filter.
\end{abstract}

\keywords{research idea generation \and knowledge graph \and categorical clustering
  \and cross-domain analogy \and LLM plausibility judge \and falsifier hypothesis}

\begin{figure}[t]
\centering
\includegraphics[width=0.98\linewidth]{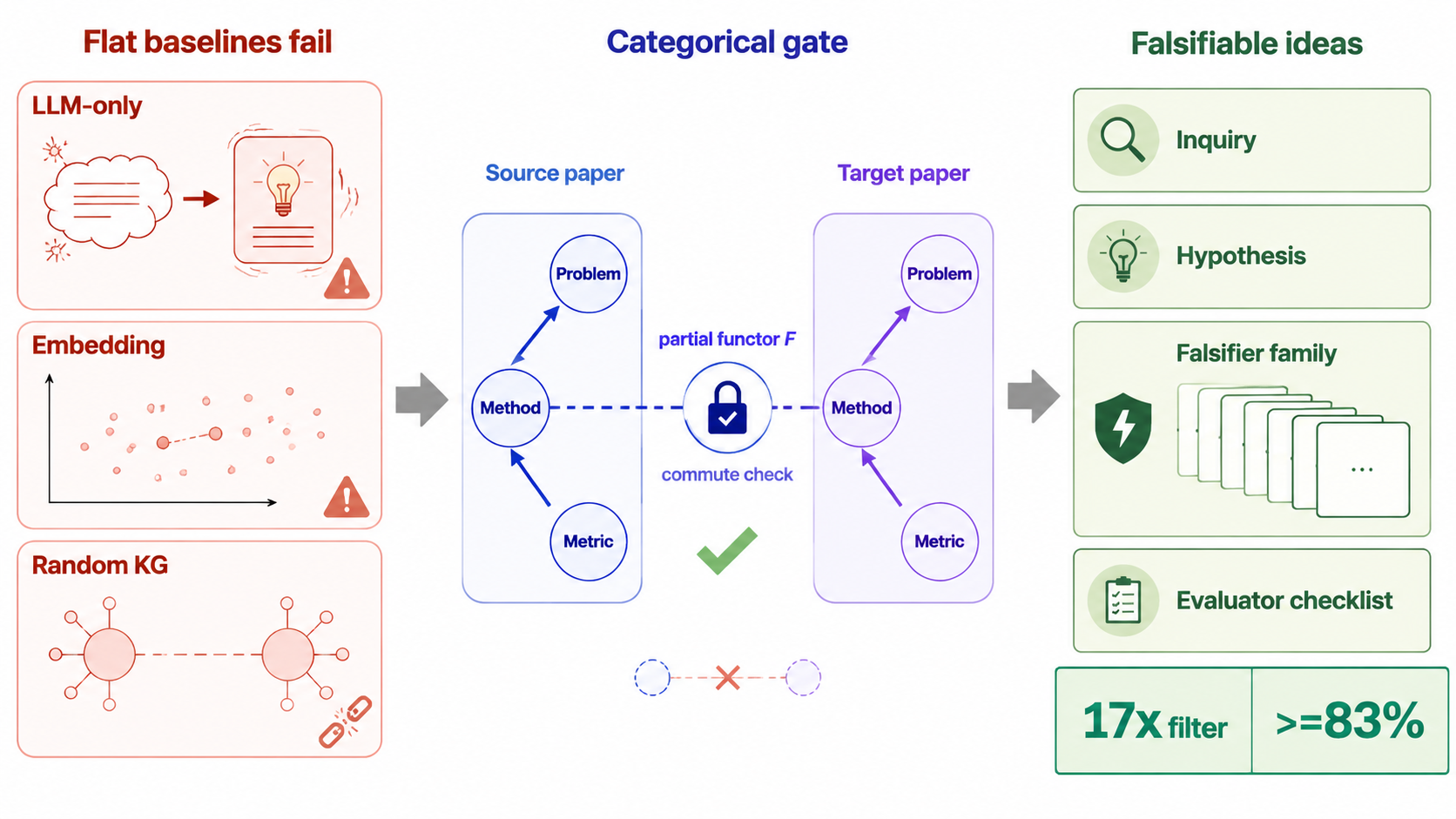}
\caption{Overview of the proposed categorical front end for research
idea generation. \emph{Left:} flat ideation baselines operate on text,
vectors, or untyped node pairs and therefore lose the typed arrows that
make an analogy checkable. \emph{Center:} the proposed front end treats
each paper as a small categorical structure and screens candidate
cross-domain bridges with a partial-functor / commutativity-style gate.
\emph{Right:} surviving bridges are converted into falsifiable idea
families with explicit evaluator-facing tests.}
\label{fig:teaser}
\end{figure}

\section{Introduction}
\label{sec:intro}

A growing class of ``AI scientist'' systems automate the full research loop
-- literature review, ideation, code, experiments, and write-up -- using
large language models (LLMs) as the active
component~\cite{Yamada2025AIScientistV2, Schmidgall2025AgentLab, Gottweis2025Kosmos, Tang2025AIResearcher}.
A weakness shared by these systems is the \emph{idea-generation} front end:
the search for promising research directions is typically reduced to free-text
prompting, LLM-as-judge novelty scoring, or retrieval over a flat embedding
of paper abstracts. Empirical evaluations are unkind to this design. A
100-author human study reports that LLM-generated ideas are rated
\emph{more} novel than expert ones but \emph{less} feasible~\cite{Si2024HumanStudy};
SoundnessBench~\cite{Wang2026SoundnessBench} concludes that ``current LLMs
are not yet reliable as stand-alone first-gate evaluators for scientific
rigor.''

We argue in this paper that the failure is not a tuning problem -- it is a
\emph{representational} one. LLM-only recombination, random paper pairing,
and embedding-similarity retrieval all share a single weakness: they treat
each paper as a flat object (a text blob or a single vector) and therefore
cannot represent the internal structure that makes a paper a \emph{paper}:
a problem, a method, a metric, a dataset, a claim, and the arrows asserting
which method solves which problem and which metric quantifies which claim.
When a researcher looks for a real cross-domain analogy, the move is not
``find a paper whose abstract is nearby in some embedding''; it is ``find
a paper whose internal arrows can be mapped, one-to-one, onto the arrows of
mine, so that the diagram still commutes.'' The first three approaches
cannot ask this question -- they have no arrows to map.

\paragraph{From bridge requirements to categorical constraints.}
A valid cross-paper bridge is useful only when its structural commitments
are explicit and checkable. The task itself imposes three requirements on
such a bridge. First, the bridge must preserve \emph{typed units}:
a method should map to a method, a metric to a metric, and a problem to a
problem. Second, it must preserve \emph{typed relations}: if the source
paper asserts that a method solves a problem or that a metric quantifies a
claim, the target side must contain the corresponding relation rather
than merely the same vocabulary. Third, these preserved relations must be
coherent when chained: transferring ``method solves problem'' and
``metric evaluates method'' should still yield a target-side mechanism
that can be checked. These are exactly the minimal ingredients of a small
category -- objects, morphisms, composition, identity arrows, and maps
that preserve this structure.
Thus a candidate cross-domain idea becomes a partial structure-preserving
map, and the algorithm in \S\ref{sec:method} implements only the checks
needed for that map: signature matching, arrow-preservation filtering,
and a six-axis judge that asks whether the local square closes.

\paragraph{What this paper contributes.}
The paper has \emph{algorithmic} contributions, not infrastructural ones.
\begin{enumerate}[leftmargin=*,topsep=2pt,itemsep=1pt]
  \item A formal model of paper-level structure as a category and of
    research-idea generation, including cross-domain bridges, as a
    partial-functor search problem
    (\S\ref{sec:why}--\ref{sec:catmodel}).
  \item A concrete algorithm that operationalises the model in three
    layers: (i) a \emph{categorical signature} schema that lifts
    typed \emph{object$\times$metric} pairs to typed morphisms with a
    de-boilerplate gate against universally-frequent signatures;
    (ii) a sparse paper$\times$paper graph clustered into 216
    intra-discipline communities; (iii) a functor-preservation gate
    plus a six-axis LLM \emph{plausibility judge} that operationalises
    the commutativity question with explicit per-axis rationales
    (\S\ref{sec:method}).
  \item An evaluation under four experimental conditions
    -- within$\times$cross $\times$ dump$\times$complete -- that ablate
    the contribution of each algorithm layer
    (\S\ref{sec:setup}--\ref{sec:exp}).
\end{enumerate}

The categorical KG we use as substrate is built by our prior
work, PARNESS~\cite{Wang2025Parness}, which provides full-text PDF
parsing, typed-entity extraction, and the underlying knowledge-graph
storage. The data layer is not a contribution of this paper; the
algorithm that runs on top of it is.

\paragraph{Position vs the sheaf-obstruction wave.}
Applied category theory has produced a recognisable 2025--2026 wave that
uses sheaf cohomological obstruction as a coherence detector for
distributed AI beliefs: HOLOGRAPH~\cite{Kim2025Holograph} reads
non-vanishing $H^1$ of LLM-prior presheaves as obstruction to global
causal structure; Sheaf-Laplacian
Obstruction~\cite{Sloboda2026SheafObstruction} formalises cross-modal
alignment hardness; Olivieri--Hern\'andez detect AI-agent theory-shift
by checking transport and gluing failure on a finite
sheaf~\cite{Olivieri2026TheoryShift}. None of these targets idea
generation. We adopt the gate-level intuition -- ``is this analogy a
real morphism or a homonymy artefact?'' -- and apply it at the
cross-paper level, with the LLM operationalising commutativity instead
of $H^1$.

\section{Why Existing Approaches Are Insufficient}
\label{sec:why}

Before describing our algorithm we explain why three standard approaches
to research idea generation do not solve the problem we care about. The
three are: (B1) LLM-only ideation; (B2) random pairing over an existing
paper KG; and (B3) embedding-similarity retrieval. We treat them as
\emph{baselines for the representational question}: each makes a
different assumption about how to represent a paper, and each fails for
a corresponding reason.

\paragraph{B1: LLM-only recombination.}
A modern LLM prompted with ``generate $N$ cross-disciplinary research
ideas'' produces fluent, surface-novel text. Three failure modes are
documented in the literature and observed in our own pilots. First,
the LLM hallucinates citations: an idea typically arrives with
``inspired by Smith et al. 2022'' that does not exist or does not say
what it is claimed to say~\cite{Yamada2025AIScientistV2, Wang2026SoundnessBench}.
Second, the LLM has no notion of a \emph{morphism}: an
idea of the form ``apply $X$ from $A$ to $B$'' is generated by string
substitution, without checking that $X$'s domain in $A$ even type-checks
in $B$. Third, the LLM's distribution of mechanisms collapses --
\emph{mode collapse} -- toward a handful of fashionable
techniques (diffusion, attention, contrastive learning), regardless
of source paper~\cite{Unlocking2026AR}.

\paragraph{B2: Random paper pairing on a KG.}
Given a paper KG, one could pick two papers uniformly at random and ask
an LLM to ideate a bridge. This is what an unconstrained
``pair-and-prompt'' baseline does. The combinatorial space is too large
to be useful: on our corpus of 17{,}650 full-text-parsed papers
(\S\ref{sec:data}),
$\binom{17650}{2}\approx 1.56\!\times\!10^{8}$ pairs are available; the
overwhelming majority of randomly selected pairs share no plausible
mechanism, and the LLM's output is forced to either refuse or fabricate
one. There is no signal to filter on -- random pairing assumes that a
functor between the two papers exists rather than checking.

\paragraph{B3: Embedding-similarity retrieval.}
The most common research-aware approach is to embed each paper's
abstract (or its typed entities) in a dense vector space and retrieve
nearest neighbours~\cite{Wang2023SciMON, Liu2025KGDesignByAnalogy}. This
is strictly stronger than B1/B2 but suffers from a specific failure
mode that we observed repeatedly in our judge rationales:
\emph{homonymy collapse}. Two papers with high embedding similarity
often share \emph{vocabulary} without sharing \emph{structure}: the
word ``mechanism'' in a circuit netlist (DE-HNN) means a solver step,
and in equivariant image registration (CARL) means an equivariant
feature transformation -- two completely different morphisms with the
same surface token. A cosine distance cannot tell these apart. More
generally, an embedding is a function $\mathcal{C}_p \to \mathbb{R}^d$
that flattens the entire internal structure of paper $p$ into a single
vector; the morphisms of $\mathcal{C}_p$ are quotiented out. What
embedding similarity surfaces is exactly the set of pairs for which
this quotient agrees -- which is necessary but far from sufficient for
a genuine cross-domain analogy.

\paragraph{The shared failure.}
The three baselines fail for one reason: they represent each paper as a
flat object (string, vector, or KG node without typed internal arrows).
A researcher looking for a real cross-domain analogy does not work this
way. Table~\ref{tab:baseline_failure_summary} summarises the contrast:
different surface implementations, same structural loss. The next section
makes that argument precise.

\begin{table}[t]
\centering
\caption{Why flat ideation baselines fail, and what the categorical
front end restores. The point is representational: without typed arrows,
composition, and a bridge-existence check, a generator can produce fluent
ideas but cannot tell whether the proposed analogy is structurally valid.}
\label{tab:baseline_failure_summary}
\small
\begin{tabularx}{\linewidth}{@{}l l X X X@{}}
\toprule
\textbf{Method} & \textbf{Representation} & \textbf{What it loses} & \textbf{Failure mode} & \textbf{Our fix} \\
\midrule
LLM-only & text & typed arrows & hallucination; type mismatch & morphism-grounded prompt plus judge \\
Embedding retrieval & vector & composition and commutativity & homonymy collapse & partial-functor gate \\
Random KG pairing & node pair & bridge-existence check & fabricated bridge & preservation-rate filter \\
Categorical KG & objects, morphisms, composition & preserves structure & checkable analogy & commute-style gate \\
\bottomrule
\end{tabularx}
\end{table}
\section{A Categorical Model of Research-Paper Structure}
\label{sec:catmodel}

This section formalises the structural requirements identified above.
The presentation follows the checks a cross-paper idea must pass and then
introduces the categorical vocabulary needed to state those checks
precisely.
The correspondence to concrete system components appears in
Table~\ref{tab:math_system}.

\paragraph{What a valid bridge must preserve.}
Suppose paper $p$ contributes a method $M_p$ that solves problem $P_p$
under metric $m_p$, and we want to transfer that mechanism into a target
paper $q$. A valid bridge cannot be only a nearby abstract, a shared
keyword, or a plausible sentence. It must answer three concrete questions:
\begin{enumerate}[leftmargin=*,topsep=2pt,itemsep=1pt]
  \item \textit{Are the units well typed?} The image of a method should be
    a method-like unit in $q$, not a dataset or a generic theme.
  \item \textit{Are the relations preserved?} If $p$ asserts
    $M_p \xrightarrow{\mathrm{solves}} P_p$, the proposed image in $q$
    must assert the analogous typed relation, not merely share a word such
    as ``mechanism'' or ``alignment.''
  \item \textit{Do relation chains remain coherent?} When relations are
    composed -- for example a method is evaluated by a metric that
    quantifies a claim -- the transferred chain should still describe a
    target-side mechanism that can be checked.
\end{enumerate}
A typed KG can express the first two questions. The third requires a
notion of composition, and the bridge must preserve that composition.
That is the precise point at which the model becomes categorical.

\paragraph{The minimal formal object.}
We use only the elementary vocabulary of category theory. A
\emph{category} $\mathcal{C}$ consists of objects, morphisms between
objects, identities, and an associative composition operation
\begin{equation}
\circ\colon\ \mathrm{Mor}(B,C) \times \mathrm{Mor}(A,B)\ \longrightarrow\ \mathrm{Mor}(A,C).
\label{eq:composition}
\end{equation}
A \emph{functor} $F\colon \mathcal{C}\to\mathcal{D}$ maps objects and
morphisms while preserving identities and composition:
\begin{equation}
F(g\circ f) \;=\; F(g)\circ F(f),
\qquad
F(\mathrm{id}_A) \;=\; \mathrm{id}_{F(A)}.
\label{eq:functor-laws}
\end{equation}
A \emph{partial functor} $F\colon \mathcal{C}\rightharpoonup\mathcal{D}$
is the same structure-preserving map but defined only on a subcategory;
this is the right object for research ideation because an analogy usually
transfers only a mechanism, not an entire paper. A diagram commutes when
two directed paths with the same endpoints compose to the same morphism.
For example, if one path from $A$ to $B$ has arrows
$f_1,\ldots,f_k$ and the other has arrows $g_1,\ldots,g_r$, then
\begin{equation}
f_k \circ \cdots \circ f_1 \;=\; g_r \circ \cdots \circ g_1
\quad \in \mathrm{Mor}(A,B).
\label{eq:commute}
\end{equation}
For standard accounts see Mac Lane~\cite{MacLane1971CT},
Awodey~\cite{Awodey2006CT}, Spivak~\cite{Spivak2014CTSci}, and
Fong--Spivak~\cite{FongSpivak2019AppliedCT}.

\paragraph{Paper structure as a small category.}
For each paper $p$, we instantiate a small category $\mathcal{C}_p$.
Its objects are the extracted entity instances in the paper, each with a
kind map
\begin{equation}
\tau_p\colon \mathrm{Ob}(\mathcal{C}_p) \longrightarrow
\bigl\{\textsc{Problem},\,\textsc{Method},\,
  \textsc{Metric},\,\textsc{Dataset},\,\textsc{Claim},\ldots\bigr\}.
\label{eq:objs}
\end{equation}
Its morphisms are the typed relations asserted between those entity
instances:
\begin{equation}
\textsc{Method} \xrightarrow{\;\mathrm{solves}\;} \textsc{Problem},
\quad
\textsc{Metric} \xrightarrow{\;\mathrm{quantifies}\;} \textsc{Claim},
\quad
\textsc{Dataset} \xrightarrow{\;\mathrm{evaluates}\;} \textsc{Method}.
\label{eq:morphs}
\end{equation}
This follows the applied-category-theory view of scientific schemas as
small categories, close to Spivak's olog-style modelling of typed
concepts and asserted relationships~\cite{Spivak2014CTSci}.

\paragraph{The corpus-scale proxy: categorical signatures.}
The full $\mathcal{C}_p$ is too large and noisy to compare exhaustively
for every pair of papers. The algorithm therefore manipulates a compact
signature,
\begin{equation}
\sigma_p \;=\; \bigl\{\,(\textsc{ObjKind},\,\textsc{MetricKind})\,\bigr\}_p,
\label{eq:sigma}
\end{equation}
the family of typed object--metric signature atoms extracted from
paper $p$'s method entities after boilerplate demotion. Each atom is a
projection of one or more paper-internal morphisms to their typed
source/target-and-metric information, not an arbitrary keyword. Shared
signature elements retrieve paper pairs whose internal arrows have a
candidate match, while the later gates decide whether that match is a
real bridge or a homonymy artefact.

\begin{figure}[t]
\centering
\begin{tikzpicture}[
  font=\small,
  obj/.style={draw, rounded corners=2pt, minimum width=1.3cm,
              minimum height=0.65cm, inner sep=2pt, fill=white},
  morph/.style={->, >=Stealth, thick},
  funct/.style={->, >=Stealth, dashed, thick, blue!55!black}
]
\node[obj] (Mp) at (0, 2)   {$M_p$};
\node[obj] (Pp) at (4, 2)   {$P_p$};
\draw[morph] (Mp) -- node[above, font=\scriptsize]{\textsc{solves}} (Pp);
\node[left=2pt of Mp, font=\scriptsize, gray] {$\mathcal{C}_p$};

\node[obj] (Mq) at (0, 0)   {$M_q$};
\node[obj] (Pq) at (4, 0)   {$P_q$};
\draw[morph] (Mq) -- node[below, font=\scriptsize]{\textsc{solves}} (Pq);
\node[left=2pt of Mq, font=\scriptsize, gray] {$\mathcal{C}_q$};

\draw[funct] (Mp) -- node[left,  font=\scriptsize]{$F$} (Mq);
\draw[funct] (Pp) -- node[right, font=\scriptsize]{$F$} (Pq);

\node[font=\scriptsize, gray, align=left] at (7.2, 1)
  {source relation maps to\\target relation};
\end{tikzpicture}
\caption{The commute condition the plausibility judge of this paper
operationalises. A partial functor
$F\colon \mathcal{C}_p \rightharpoonup \mathcal{C}_q$ (dashed blue) is a
valid local bridge when applying $F$ to the source-paper arrow
\textsc{method-solves-problem} yields the corresponding target-paper
arrow. The embedding-similarity baseline (B3 of the
preceding section) can match the four \emph{objects} pairwise but cannot
certify that the \emph{square closes}. Of the six axes the plausibility
judge scores, the two highest-weight ones --
\emph{type agreement} (do the mathematical types of the source
arrow and its image agree?) and \emph{vocabulary genuineness} (is the
matching vocabulary a real morphism rather than a homonymy artefact?)
-- are exactly what asks whether the square closes.}
\label{fig:commute}
\end{figure}
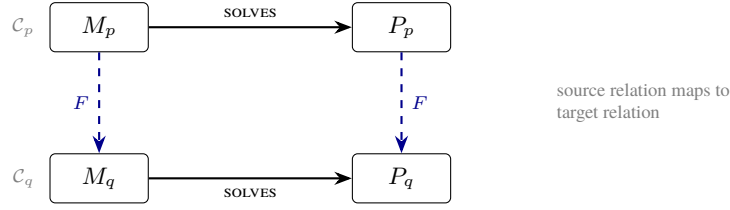

\paragraph{A candidate idea as a partial structure-preserving map.}
A cross-domain bridge from paper $p$ to paper $q$ is represented as a
candidate partial functor
\begin{equation}
F\colon \mathcal{C}_p \rightharpoonup \mathcal{C}_q.
\label{eq:F}
\end{equation}
In system terms, this means a paper pair together with matched typed
objects, matched signature elements, and evidence pointers for the source
and target relations. It must preserve object kinds, map covered source
relations to target relations of the corresponding kind, and satisfy a
local commutativity check. The last condition is the formal version of
``does the transferred mechanism still work?'' Figure~\ref{fig:commute}
shows the smallest instance.

\paragraph{What is actually checked.}
The implementation does not enumerate all paths in $\mathcal{C}_p$ and
does not run a theorem prover. Instead, it uses a scalable local
approximation to the functor and commutativity requirements. Object-kind
agreement checks whether $F$ is well typed. Morphism-class agreement and
preservation rate check whether source arrows are mapped to target arrows
of the same kind. The six-axis plausibility judge then asks whether the
proposed transfer is a real mechanism rather than lexical coincidence:
\emph{type agreement}, \emph{vocabulary genuineness}, and
\emph{mechanism specificity} are the direct local checks, with the
remaining axes recording evidence quality, bridgeability, and novelty.
Thus every accepted idea has a categorical witness in the operational
sense used here: a typed source arrow, a typed target arrow, and an
explicit judged claim that the local square closes.

\paragraph{Why this is more than a typed graph.}
A labelled KG supplies nodes and edges, so it can represent typed units
and typed relations. What it lacks by itself is the preservation question:
when a bridge maps a source relation chain into a target relation chain,
should the two paths be considered the same transferred mechanism? Category
theory adds exactly the required language -- composition, identity,
functorial preservation, and commutativity -- without requiring heavier
machinery. In this paper those notions act as an engineering
specification: they determine the schema, the candidate miner, the
preservation-rate gate, and the judge rubric.

\paragraph{Math-to-system correspondence.}
Table~\ref{tab:math_system} is the contract between the structural
requirements and the implementation. Each row names a mathematical role,
the concrete artefact that carries it, and the component that implements
or checks it. The algorithm in \S\ref{sec:method} is simply this contract
made executable at corpus scale.

\begin{table}[t]
\centering
\caption{Math$\leftrightarrow$system correspondence. The categorical
construction defined above is the specification; each row names which
concrete component implements that piece, and where in the algorithm
description and data layer it is realised.}
\label{tab:math_system}
\small
\begin{tabularx}{\linewidth}{@{}>{\raggedright\arraybackslash}p{0.26\linewidth} >{\raggedright\arraybackslash}X >{\raggedright\arraybackslash}p{0.25\linewidth}@{}}
\toprule
\textbf{Mathematical object} & \textbf{In the system} & \textbf{Implemented at} \\
\midrule
Category $\mathcal{C}_p$ (paper $p$)
  & paper $p$'s typed-entity sub-graph
  & PARNESS extractor \\
$\mathrm{Ob}(\mathcal{C}_p)$ -- entity instances
  & extracted entities with kind map $\tau_p$ to \textsc{Problem}, \textsc{Method}, \textsc{Metric}, \textsc{Dataset}, \textsc{Claim}, \ldots
  & \code{ikg\_entity.id} plus\newline \code{ikg\_entity.kind} \\
$\mathrm{Mor}(\mathcal{C}_p)$ -- typed arrows
  & relations (\textsc{solves}, \textsc{quantifies}, \textsc{evaluates}, \ldots)
  & paper-internal relation table \\
Signature atom $(\textsc{ObjKind}, \textsc{MetricKind})$
  & projection of covered morphisms into the signature family $\sigma_p$
  & typed-pair schema; cf.\ Fig.~\ref{fig:schema_algo} \\
Composition $g\circ f$ in $\mathcal{C}_p$
  & chained relations within paper $p$ (e.g.\ method-of-method evaluated by metric)
  & relation chain in the entity graph \\
Identity $\mathrm{id}_A$
  & trivial self-relation
  & implicit \\
Corpus-level community
  & community of papers sharing $\sigma_p$ structure
  & \code{paper\_\allowbreak categorical\_\allowbreak clusterer} (Leiden / CPM) \\
Partial functor $F\colon \mathcal{C}_p \rightharpoonup \mathcal{C}_q$
  & candidate cross-domain bridge
  & \code{cross\_paper\_\allowbreak candidate\_\allowbreak finder} \\
Object-type preservation by $F$
  & same-kind / different-paper filter on entity pairs
  & nearest-neighbour same-kind filter inside the cross-domain miner \\
Arrow preservation by $F$
  & \emph{preservation rate} on candidate relations
  & \code{functor\_\allowbreak preservation\_\allowbreak gate} \\
Commutativity of a small square
  & ``does the analogy actually close?''
  & six-axis plausibility judge (Fig.~\ref{fig:judge}) \\
Non-commuting square (obstruction)
  & judge verdict \textsc{no\_bridge} / \textsc{speculative}
  & advisory rejection with archived rationale \\
\bottomrule
\end{tabularx}
\end{table}

\section{Related Work}
\label{sec:related}

We position our work against four threads. The first is the
foundation on which the data substrate is built; the others are
contemporary work in the same problem space.

\paragraph{Foundation: PARNESS.}
The categorical paper KG used in this work is built by
PARNESS~\cite{Wang2025Parness}, our prior work introducing a DAG-kernel
harness for end-to-end automated research. PARNESS provides full-text
PDF parsing (with graceful abstract-only fallback for papers where
legal full-text access is unavailable), a typed-entity extractor for
turning paper bodies into typed-entity nodes, and the
underlying knowledge-graph storage. The present paper sits one layer
above: PARNESS supplies the typed-entity graph; we contribute the
categorical signature schema, the clustering algorithm, the
functor-preservation gate, and the plausibility judge that run on
top.

\paragraph{End-to-end AI scientists.}
Sakana v2~\cite{Yamada2025AIScientistV2} introduced agentic tree search and
produced an entirely AI-generated workshop paper.
Agent Laboratory~\cite{Schmidgall2025AgentLab} demonstrates that the
literature/experiment/report loop runs at 84\% lower cost than its
predecessor. Kosmos~\cite{Gottweis2025Kosmos} extends the run length to
$\sim$12 hours of parallel data analysis, literature search, and
hypothesis generation. AI-Researcher~\cite{Tang2025AIResearcher} stresses
``systematic exploration beyond cognitive limitations.''
ResearchAgent~\cite{Baek2024ResearchAgent} couples idea proposal with
collaborative reviewing agents. All of these systems accept an idea seed
from the user or use free-text LLM judgment to grade their own ideas.
Independent audits of the earlier Sakana system note that literature
retrieval is weak and that ``novelty often repeats existing work,'' a
problem that motivates our structural front end.

\paragraph{Structured / KG-grounded idea generation.}
SciMON~\cite{Wang2023SciMON} introduced literature-grounded ``inspirations''
with an explicit novelty objective. ResearchAgent and
Chain-of-Ideas~\cite{Li2024ChainOfIdeas} organise the literature into linear
research chains; VirSci~\cite{Su2024VirSci} uses heterogeneous expert agents.
Scideator~\cite{Radensky2024Scideator} extracts and recombines
(purpose, mechanism, evaluation) facets; CHIMERA~\cite{Sternlicht2025Chimera}
mines recombination instances at scale as an IE task.
KG-CoI~\cite{Xiong2024KGCoI} anchors reasoning in an external KG to reduce
hallucination. Knowledge graph-assisted design-by-analogy~\cite{Liu2025KGDesignByAnalogy}
formalises function-effect-structure ontologies. Nova~\cite{Wang2024Nova}
uses iterative planning and retrieval to boost diversity; IRIS~\cite{Sahu2025IRIS}
adds MCTS test-time compute and human-in-the-loop. LacMaterial~\cite{LacMaterial2025}
makes the analogical move explicit (``planet:sun :: electron:nucleus'') in
materials discovery. Most of these systems use similarity-based or
LLM-judge filters; none combines a categorical clustering of the paper KG
with a multi-axis structural gate.

\paragraph{Categorical and sheaf-theoretic gates.}
The sheaf-on-KG line begins with Knowledge
Sheaves~\cite{Hansen2021KnowledgeSheaves} (KG embedding as approximate
global section of a cellular sheaf) and Sheaf Neural
Networks~\cite{Hansen2020SheafNN, Barbero2022SheafNN}.
Boudourides~\cite{Boudourides2026SheafSemantics} extends this to a
Grothendieck topology on the free category of a KG.
\emph{Categorical Deep Learning}~\cite{Gavranovic2024CategoricalDL} argues
for monads in 2-categories of parametric maps as a unifying language for
neural architectures. A separate 2025--2026 wave applies sheaf
\emph{cohomology} as an obstruction detector for AI coherence:
HOLOGRAPH~\cite{Kim2025Holograph} for causal discovery with LLM priors,
Sheaf-Laplacian Obstruction~\cite{Sloboda2026SheafObstruction} for
cross-modal compatibility, Causal Abstraction
Networks~\cite{DAcunto2025CausalAbstractions} for multi-agent causal
beliefs, and Olivieri--Hern\'andez~\cite{Olivieri2026TheoryShift} for
detecting agent theory-shift via gluing failure. None applies the
obstruction-as-gate pattern to cross-paper research idea generation
grounded in a real paper KG. Analogous recent work,
\emph{Unlocking LLM Creativity through Analogical
Reasoning}~\cite{Unlocking2026AR}, motivates the same goal -- structural
analogy as a brake on mode collapse -- without category-theoretic
formalism. The cognitive-science roots of this move trace to Gentner's
structure-mapping theory~\cite{Gentner1983StructureMapping}.

\paragraph{Evaluation.}
Si et al.~\cite{Si2024HumanStudy} run the first head-to-head human
evaluation of LLM ideation. IdeaBench~\cite{Guo2025IdeaBench} and AI Idea
Bench 2025~\cite{Qiu2025AIIdeaBench2025} provide standardised idea-quality
benchmarks. MLR-Bench~\cite{Wang2025MLRBench} evaluates open-ended ML
research agents end-to-end. SoundnessBench~\cite{Wang2026SoundnessBench}
specifically targets the first-gate soundness question and concludes that
LLMs are unreliable as the sole soundness grader -- a finding that
shapes our decision to operate the judge as advisory (fail-open) rather
than blocking.

\section{Data and KG Construction}
\label{sec:data}

The corpus and KG that our algorithm runs on are produced by
PARNESS~\cite{Wang2025Parness}; we do not claim them as a contribution.
This section summarises only the aspects relevant to interpreting the
experiments and to guiding the schema decisions that drive the
categorical idea-mining algorithm presented in
\S\ref{sec:method}.

\paragraph{Corpus.}
The substrate this paper's algorithm runs on is a curated set of
\textbf{17{,}650 full-text-parsed papers} -- PDF body, sections, figures,
tables and formulas extracted by the PARNESS PDF pipeline into typed
records -- together with an additional \textbf{9{,}081 abstract-only
papers} for which legal full-text retrieval is unavailable. The
abstract-only papers contribute metadata to the KG but no method-step
signatures, and are therefore invisible to the within-domain miner of
\S\ref{sec:method}. Both sets of paper IDs are released as static text
files alongside the source repository so that other groups can
replicate or compare against the same corpus without access to the
parsed data itself --
the full-text list~\href{https://github.com/gtrhythm/PARNESS/blob/main/papers/categorical_kg_idea_mining/inputs/kg_papers_full_text.txt}{\texttt{kg\_papers\_full\_text.txt}}
and the abstract-only list~\href{https://github.com/gtrhythm/PARNESS/blob/main/papers/categorical_kg_idea_mining/inputs/kg_papers_abstract_only.txt}{\texttt{kg\_papers\_abstract\_only.txt}}.

\paragraph{Generated-idea artifacts.}
We also release the generated opportunity records, judge-survivor list,
proposal records, and five-form hypothesis-family outputs as static
JSON/JSONL/CSV artifacts in the PARNESS repository:
\href{https://github.com/gtrhythm/PARNESS/tree/main/papers/categorical_kg_idea_mining/artifacts}{\texttt{papers/categorical\_kg\_idea\_mining/artifacts}}.
The artifact contains generated ideas and audit metadata only; it does
not redistribute parsed full text or copyrighted paper contents.

\paragraph{Parsing.}
Body sections, claims, methods, metrics and similar typed entities are
extracted by a structured LLM-call pipeline using
\emph{MiniMax-M2.7} as the extraction model. The PARNESS layer is what
turns each PDF into the typed-entity records (covering
\textsc{Method}, \textsc{Hypothesis}, \textsc{Observation},
\textsc{Contribution}, \textsc{Claim}, \textsc{Critique},
\textsc{Inquiry}, \textsc{Branch}, and a dozen more typed kinds) that
this paper's algorithm reads.

\paragraph{Schema choices relevant to this paper.}
Three properties of the substrate matter for the algorithm in
\S\ref{sec:method} and would be missed if we treated the KG as ``a
generic property graph''.
\begin{enumerate}[leftmargin=*,topsep=2pt,itemsep=2pt]
  \item \emph{Object--metric typing is required, not optional.}
    Each \textsc{Method} entity carries an explicit pair recording the
    \emph{kind} of object the method targets and the \emph{kind} of
    metric used to evaluate it. These pairs are the smallest unit on
    which our \emph{categorical signatures} (\S\ref{sec:catmodel}) can
    be defined; without typed extraction the morphism family $\sigma_p$
    collapses into a bag of keywords and the homonymy problem of
    \S\ref{sec:why} returns.
  \item \emph{Section-level provenance is retained.}
    Every typed entity points back to the paragraph and section that
    produced it. This is what makes the \emph{evidence pointer} on a
    generated idea verifiable by a human reader, and is what allows the
    de-boilerplate gate (next item) to be evaluated correctly.
  \item \emph{Universally-frequent signatures are demoted, not dropped at
    extraction time.} Signatures with document frequency $\geq$80\%
    (typically ``release code,'' ``comprehensive evaluation,'' and
    similar) are flagged \texttt{boilerplate=true} but left in the
    record. The algorithm in \S\ref{sec:method} can then choose to
    ignore them at clustering time while still using them for
    completeness when answering an evidence query.
\end{enumerate}
None of these properties is specific to the underlying storage engine;
we treat the KG as an abstract typed graph and do not depend on the
storage details. Figure~\ref{fig:schema_algo} makes the link between
schema choices and algorithm operations explicit.

\begin{figure}[t]
\centering
\begin{tikzpicture}[
  font=\small,
  schemabox/.style={draw, rounded corners=3pt, fill=blue!5,
                    minimum width=5.0cm, minimum height=1.0cm,
                    align=center, text width=4.7cm, inner sep=4pt},
  algobox/.style={draw, rounded corners=3pt, fill=green!7,
                  minimum width=5.0cm, minimum height=1.0cm,
                  align=center, text width=4.7cm, inner sep=4pt},
  flow/.style={->, >=Stealth, thick, gray!70!black},
  hdr/.style={font=\small\bfseries}
]
\node[hdr] at (-3.0, 4.4) {Data-layer schema choice};
\node[hdr] at ( 3.0, 4.4) {Algorithm operation};

\node[schemabox] (s1) at (-3.0, 3.1)
  {\textbf{(1)} typed \emph{object-kind} $\times$ \emph{metric-kind}
   pairs on every \textsc{Method} entity};
\node[algobox]   (a1) at ( 3.0, 3.1)
  {morphism family $\sigma_p$ for paper $p$\\(signature extraction)};
\draw[flow] (s1) -- (a1);

\node[schemabox] (s2) at (-3.0, 1.5)
  {\textbf{(2)} section-level provenance pointer on every typed entity};
\node[algobox]   (a2) at ( 3.0, 1.5)
  {verifiable evidence pointer on each generated hypothesis (direction proposer)};
\draw[flow] (s2) -- (a2);

\node[schemabox] (s3) at (-3.0, -0.1)
  {\textbf{(3)} boilerplate flag on universal-frequency signatures
   ($\mathrm{df}/N \geq 0.80$)};
\node[algobox]   (a3) at ( 3.0, -0.1)
  {de-boilerplate gate; clustering edge weight\\$|\sigma_p \cap \sigma_q|_{\text{non-bp}}$};
\draw[flow] (s3) -- (a3);

\node[font=\scriptsize, gray, align=center] at (0, -1.4)
  {Data-layer schema choices are dictated by what the categorical
   idea-mining algorithm needs to compute --\\not by storage
   convenience. Each supports the local preservation checks of
   Fig.~\ref{fig:commute} at corpus scale.};
\end{tikzpicture}
\caption{Schema-to-algorithm mapping. Each of the three
data-layer decisions described in the Data section is dictated by a
specific operation the categorical idea-mining algorithm of the next
section needs to perform on the signature family $\sigma_p$.
(1)~Typed object$\times$metric pairs are the \emph{atoms} of $\sigma_p$,
without which clustering would degenerate to bag-of-tokens.
(2)~Section-level provenance is what lets the evidence pointer attached
to a generated hypothesis be human-verifiable, closing the loop between
an LLM proposal and the source paper. (3)~The boilerplate flag,
demoted-but-retained, preserves completeness for retrieval while
allowing the categorical clustering to use only signature atoms that
carry IDF signal.}
\label{fig:schema_algo}
\end{figure}

\section{Method: The Categorical Idea-Mining Algorithm}
\label{sec:method}

\begin{figure}[t]
\centering
\includegraphics[width=\linewidth]{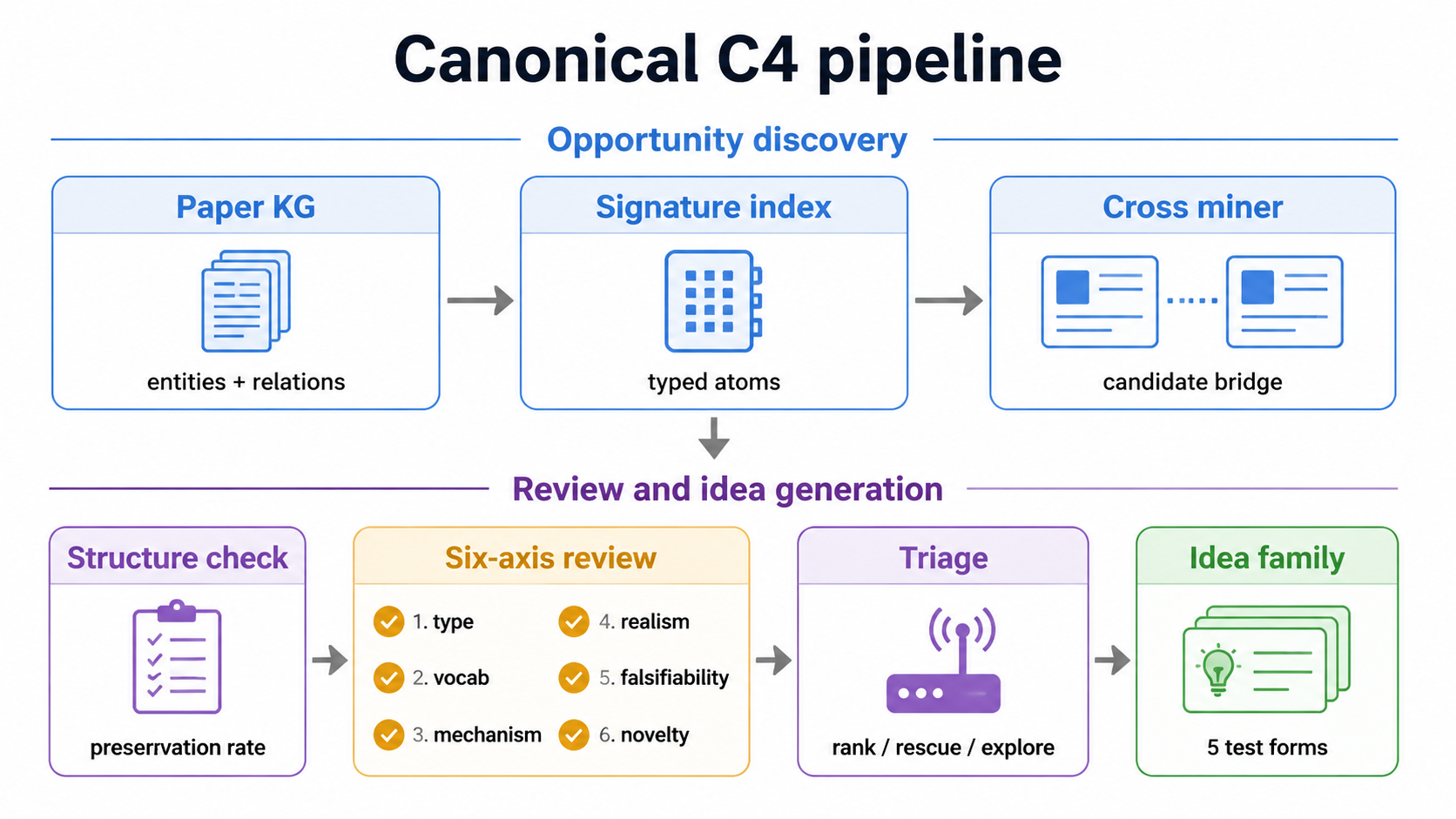}
\caption{Canonical cross-domain complete path (C4). The top row
shows opportunity discovery: the PARNESS KG supplies extracted paper
entities and relation evidence; the categorical signature index projects
each paper into non-boilerplate typed signature atoms; the cross-domain
miner uses same-kind nearest-neighbour anchors to propose candidate
bridges. The bottom row shows validation and generation: the
functor-preservation gate computes a preservation-rate signal, the
six-axis judge is applied only in the judged C4 path, triage ranks or
rescues surviving opportunities, and generation runs proposer,
reranker, and five-form family expansion. The other experimental
conditions remove or replace pieces of this path as specified in
Fig.~\ref{fig:conditions}; arrows here are system data-flow arrows, not
categorical morphisms.}
\label{fig:arch}
\end{figure}

Conceptually, the algorithm has three structural layers: (i) a
categorical signature schema that projects each paper's typed relations
into a compact signature family; (ii) a clustering step that turns the
corpus into an intra-discipline community structure; (iii) a
mining-and-judging step that enumerates partial-functor candidates
between papers and screens them with local preservation and
commutativity-style checks. Figure~\ref{fig:arch} expands these layers
for the full cross-domain judged path (C4); the four ablation variants
that remove or replace individual components are specified separately in
\S\ref{sec:setup}.

To read Fig.~\ref{fig:arch}, an \emph{opportunity} is the record that
moves through the pipeline: it starts as a candidate paper pair or
within-cluster micro-paradigm, accumulates structural evidence and gate
scores, and may end as generated hypotheses. The \emph{signature index}
is the per-paper family $\sigma_p$ of typed signature atoms defined in
Eq.~\ref{eq:sigma}; the \emph{cross miner} is the same-kind nearest-neighbour search
that proposes candidate partial-functor bridges; the \emph{functor gate}
computes the preservation-rate signal; the \emph{judge} is the six-axis
LLM plausibility scorer; \emph{triage} ranks or rescues surviving
opportunities; and \emph{generation} denotes proposer, reranker and
five-form family expansion. C4 is the full cross-domain judged path shown
in the figure. C1/C3 replace the cross miner with the within-domain miner,
and C1/C2 omit parts of the complete judged path as detailed in
Fig.~\ref{fig:conditions}.

\begin{figure}[t]
\centering
\includegraphics[width=\linewidth]{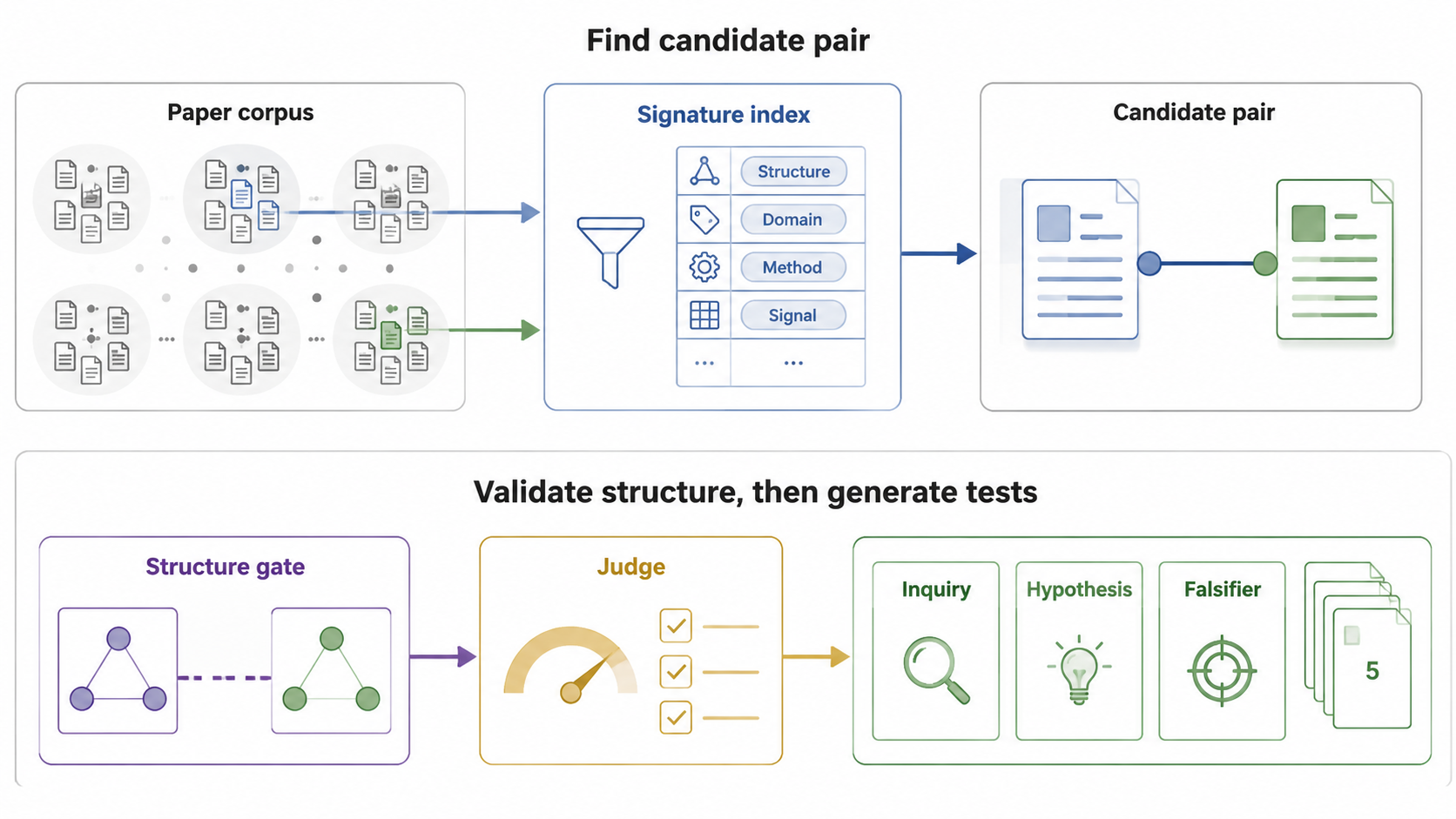}
\caption{Corpus-to-hypothesis walkthrough. The left-to-right trace
emphasizes that cross-domain opportunities are not hand-picked pairs:
papers are first indexed by typed categorical signatures, candidate
pairs are retrieved through shared structure, and the resulting bridge
is then screened by a functor-preservation gate and a plausibility
judge before being expanded into falsifiable hypothesis families. Thus
the figure illustrates both selection (\emph{which two papers should
be linked?}) and validation (\emph{does the link preserve enough
structure to be testable?}).}
\label{fig:walkthrough}
\end{figure}

Fig.~\ref{fig:arch} is the canonical path. To make it concrete,
Fig.~\ref{fig:walkthrough} shows the corresponding discovery trace:
the system first turns a large paper corpus into typed signatures,
uses shared signature atoms to retrieve candidate cross-domain paper
pairs, and only then applies the categorical preservation gate, judge
and hypothesis-family generator.

\subsection{Categorical signatures and clustering}

\paragraph{Signature extraction.}
For each paper $p$ we read its \textsc{Method} entities and emit the
signature family (cf.\ Eq.~\ref{eq:sigma})
\begin{equation}
\sigma_p \;=\; \bigl\{\,(\textsc{ObjKind}_i,\,\textsc{MetricKind}_i)\,\bigr\}_{i}.
\label{eq:sigma-extract}
\end{equation}
By the categorical model of \S\ref{sec:catmodel} each pair is a
signature atom: a typed projection of one or more morphisms in
$\mathcal{C}_p$. The family is a compact representation (typically
1--6 elements per paper after de-boilerplate) of the paper's internal
arrow structure.

\paragraph{Sparse paper$\times$paper graph.}
We construct a sparse paper$\times$paper graph whose edge weight
between $p$ and $q$ is the IDF-weighted count of shared signature
classes,
\begin{equation}
w(p,q) \;=\; \sum_{s\,\in\,\sigma_p \cap \sigma_q}\,\mathrm{idf}(s),
\qquad
\mathrm{idf}(s) \;=\; \log\!\frac{N}{1 + \mathrm{df}(s)},
\label{eq:edge-weight}
\end{equation}
where $N$ is the corpus size and $\mathrm{df}(s)$ is the document
frequency of signature class $s$. Universally-frequent signatures
($\mathrm{df}(s)/N \geq 0.80$) are dropped by the de-boilerplate gate
before the intersection is computed, and edges with $w(p,q)=0$ are
absent. Empirically the median non-boilerplate signature count per
paper is 1, so we set the edge condition to require at least one
shared non-boilerplate signature.

\paragraph{Community detection.}
We cluster the paper graph by Leiden / CPM at resolution $\gamma=1.0$
and merge sub-minimum-size communities into their strongest neighbour by
inter-community edge weight. The output is a per-paper cluster label.
On our corpus this yields 216 clusters; 76 of these participate in at
least one cross-cluster edge that survives the bridgeability prior of
\S\ref{sec:method-mining}. Operationally, communities play the role of corpus-level regions: a
cross-domain bridge connects two such regions and is witnessed by a
partial-functor candidate between two of their member papers.

\subsection{Mining: within- and cross-domain candidates}
\label{sec:method-mining}

\paragraph{Within-domain miner.}
For each cluster the within-domain miner enumerates rare-signature
sub-paradigms: tuples of signatures whose intra-cluster cohesion is
high but whose corpus-wide IDF is also high. Each surviving micro-paradigm
becomes an \emph{opportunity} tagged as a within-domain mechanism-depth
candidate, with a small set of evidence papers attached.

\paragraph{Cross-domain miner.}
The cross-domain miner is what proposes candidate partial functors. We
embed each typed-entity node into a dense space and issue a
$k$-nearest-neighbour (kNN) query ($k=5$) per anchor; we filter to
same-kind, different-paper pairs and apply a bridgeability prior
(same-cluster threshold 0.82, cross-cluster 0.80). For each surviving
pair $(p, q)$ we compute the structural \emph{preservation rate}: how
many of the candidate relations on the source paper map (under
cosine $\geq$0.70) into a relation of the same kind on the target
paper. This is the discrete local proxy for arrow preservation and the
commutativity-style check of \S\ref{sec:catmodel}. Cross-domain
opportunities are tagged as
shared-mechanism candidates and carry a list of preserved method-class
signatures.

\subsection{Gates: from candidates to bridges}

\paragraph{Functor-preservation gate.}
The functor-preservation gate reads cached preservation rates between
cluster pairs from \textsc{Functor} edges in the KG and, per opportunity,
aggregates over its parent domain pairs. Aggregated rate below
$0.30$ demotes the opportunity (score $\times 0.5$); it is operated as
a demote, not a hard block, so the downstream judge can still see the
candidate.

\paragraph{Reasonability gate.}
The reasonability gate blends three feasibility components --
\emph{method overlap} (centroid cosine on method embeddings, weight
0.50), \emph{formalism overlap} (Jaccard on the formal vocabulary,
weight 0.30), and \emph{institutional bridge} (citation proximity,
weight 0.20) -- minus a distance penalty (weight 0.30). Decisions are
tiered: $F\geq 0.60$ accepted, $0.30\leq F<0.60$ speculative, $F<0.30$
rejected.

\paragraph{Six-axis plausibility judge.}
The judge is the algorithm's operationalisation of commutativity at
the LLM level. For each opportunity an LLM is prompted to score six
axes in $[0,1]$ with explicit per-axis rationales, listed here with
their weights:
(i) \emph{type agreement} ($w=0.30$): do the mathematical types of the
source and target morphisms match?
(ii) \emph{vocabulary genuineness} ($w=0.20$): is the matching
vocabulary a real morphism rather than a homonymy artefact?
(iii) \emph{mechanism specificity} ($w=0.15$): is the morphism named
concretely, or only by an opaque label?
(iv) \emph{transferability realism} ($w=0.15$): could the source-side
mechanism plausibly be instantiated in the target setting?
(v) \emph{falsifiability within 24 hours} ($w=0.10$): can the resulting
hypothesis be checked by a short, concrete experiment?
(vi) \emph{novelty value} ($w=0.10$): does the bridge surface a
direction not already pursued in the existing literature?
The aggregate survival threshold is $0.30$. The judge is operated
\emph{advisory}: null or unparseable LLM outputs are treated as pass
(fail-open), and every rejected opportunity is archived with its full
per-axis vector and rationale for offline re-audit. We discuss the
SoundnessBench~\cite{Wang2026SoundnessBench}-driven rationale for the
advisory choice in \S\ref{sec:advisory}.

\subsection{Triage and hypothesis-family generation}

\paragraph{Triage router.}
The triage router routes surviving opportunities by a weighted blend of
judge score and structural score (weights 0.35, 0.25, 0.20, 0.10, 0.10
over preservation rate, raw rate, log-saturated number of preserved
signatures, bridgeability of physical objects, and novelty gap).
High-judge candidates land in the proceed lane; high-structure /
low-judge candidates are rescued. A small exploration quota is reserved
for under-judged but structurally rich candidates.

\paragraph{Direction proposer.}
A surviving opportunity is converted into an \textsc{Inquiry}
+ \textsc{Branch} + \textsc{Hypothesis} triple by an LLM call that is
given the canonical source/target members (centrality-filtered) and the
preserved method-class signatures. The prompt is identical across the
four experimental conditions of \S\ref{sec:setup}; what differs is
\emph{which opportunities reach it}.

\paragraph{Hypothesis-family expander.}
Each proposal is expanded into a five-form falsifier family:
\begin{itemize}[leftmargin=*,topsep=0pt,itemsep=0pt]
  \item \emph{strict:} $|m_\text{target} - m_\text{source}| / m_\text{source} \leq \varepsilon$.
  \item \emph{rank:} $m_\text{method} > m_\text{baseline}$ on the target metric.
  \item \emph{existence:} $m_\text{method} \geq m_\text{random} + \delta$.
  \item \emph{baseline\_ratio:} $m_\text{method} / m_\text{baseline} \leq \rho$.
  \item \emph{neg\_transfer:} $m_\text{method} < m_\text{random} - \delta$.
\end{itemize}
Thresholds $\varepsilon, \rho, \delta$ are filled by the LLM and
validated against a known-metric table; each form is paired with an
explicit evaluator spec.

\section{Experimental Conditions}
\label{sec:setup}

We evaluate the algorithm of \S\ref{sec:method} under four experimental
conditions arranged on two binary axes: \emph{within} vs \emph{cross}
domain (which miner is used) and \emph{dump} vs \emph{complete}
(which gates are active). These four conditions are not separate
contributions; they are an ablation that isolates the effect of the
diversifier and the plausibility judge.

\begin{figure}[t]
\centering
\begin{tikzpicture}[
  font=\small,
  cell/.style={draw, rounded corners=3pt, align=center, text width=5.4cm,
               minimum height=1.6cm, inner sep=4pt, fill=blue!4},
  hdr/.style={font=\small\bfseries},
  axislbl/.style={font=\small\itshape, gray!50!black}
]
\node[hdr] at (1.7, 2.6) {within domain};
\node[hdr] at (8.0, 2.6) {cross domain};
\node[axislbl, rotate=90] at (-2.0, 1.05) {dump};
\node[axislbl, rotate=90] at (-2.0, -1.7) {complete};

\node[cell] (c1) at (1.7, 1.05)
  {\textbf{C1} -- within $\times$ dump\\[2pt]
   \scriptsize miner $\to$ governor (annotate) $\to$ triage};
\node[cell] (c2) at (8.0, 1.05)
  {\textbf{C2} -- cross $\times$ dump\\[2pt]
   \scriptsize cross-miner $\to$ functor gate $\to$ triage};

\node[cell, fill=green!5] (c3) at (1.7, -1.7)
  {\textbf{C3} -- within $\times$ complete\\[2pt]
   \scriptsize C1 $+$ \emph{diversifier} (MMR)};
\node[cell, fill=green!5] (c4) at (8.0, -1.7)
  {\textbf{C4} -- cross $\times$ complete $+$ judge\\[2pt]
   \scriptsize C2 $+$ diversifier $+$ \emph{plausibility judge}};

\draw[thick, green!60!black, rounded corners=3pt]
  ($(c4.south west)+(-0.05,-0.05)$) rectangle ($(c4.north east)+(0.05,0.05)$);
\end{tikzpicture}
\caption{Experimental conditions as a 2$\times$2 ablation. The two axes
are the miner (within vs cross domain) and the gate configuration
(dump vs complete). Moving \emph{down} adds the diversifier; moving
\emph{down-right} additionally adds the six-axis plausibility judge.
C4 (green outline) is highlighted because it is the full cross-domain
complete setting: it keeps the cross-domain functor gate from C2 and adds
the diversifier plus the six-axis plausibility judge.}
\label{fig:conditions}
\end{figure}
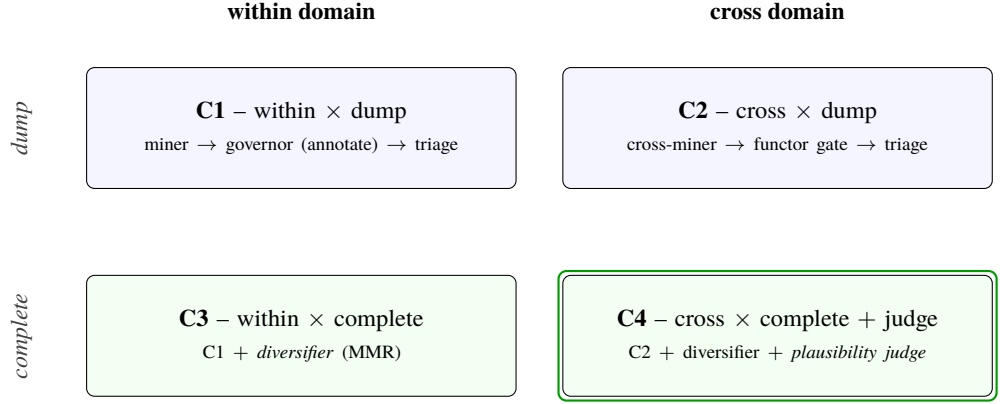

Figure~\ref{fig:conditions} lays out the 2$\times$2 ablation.

All four conditions share the same proposer prompt, the same family
expander, the same generation cap (500 proposals and 500 families per
shard, with $K=4$ shards), the same sampling parameters
(\code{temperature}$=$0.9, \code{n\_samples}$=$2), and the same upstream
KG corpus. The LLM provider is MiniMax-M2.7; the entity embedder is
qwen3-embedding-4b (2560-d).

\section{Results}
\label{sec:exp}

\begin{figure}[t]
\centering
\includegraphics[width=0.9\linewidth]{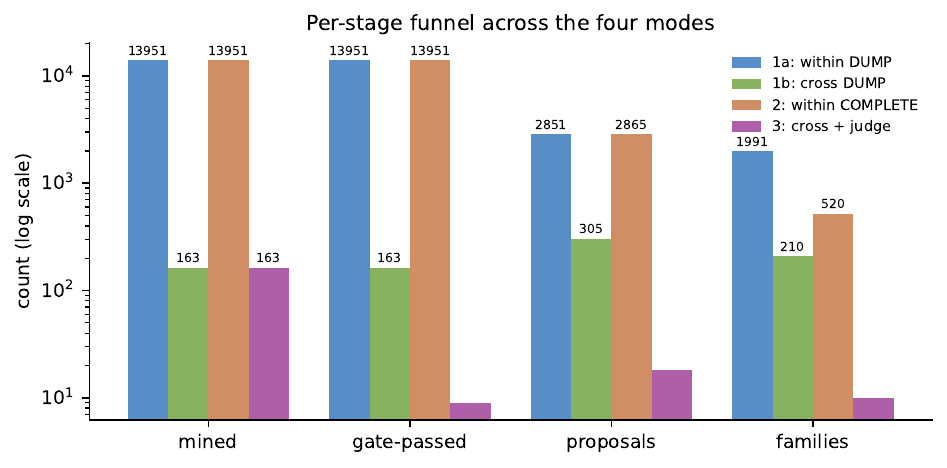}
\caption{Per-stage counts across the four conditions (log scale). The
within-vs-cross opportunity gap (86$\times$, C1/C3 vs C2/C4) and the C4
judge funnel (17$\times$) dominate the picture.}
\label{fig:funnel}
\end{figure}

\begin{figure}[t]
\centering
\includegraphics[width=0.9\linewidth]{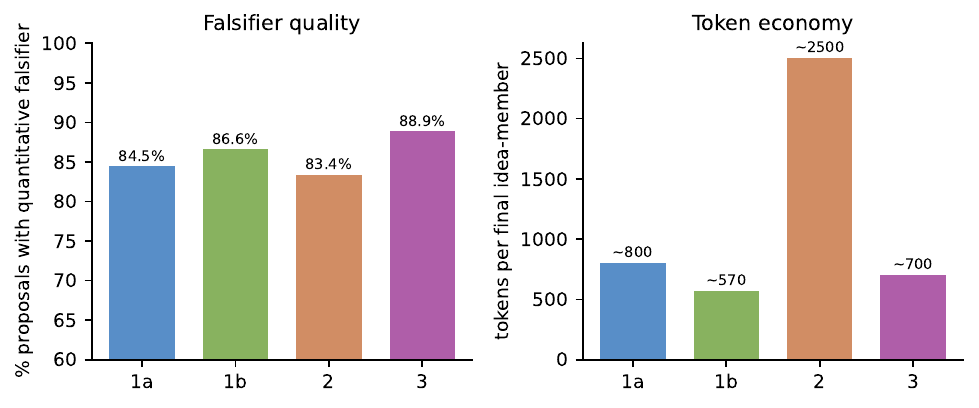}
\caption{Left: \% proposals with a quantitative falsifier (numeric
comparator + literal threshold). Right: real tokens per final idea
member. C3's outlier cost reflects transient LLM-provider retries, not a
design flaw.}
\label{fig:quality}
\end{figure}

\paragraph{Funnel.}
Table~\ref{tab:funnel} and Figure~\ref{fig:funnel} summarise the per-stage
counts. Condition C1 mines 13{,}951 within-domain opportunities; C2
mines 163 cross-domain pairs -- a 86$\times$ asymmetry. The number itself
is not the headline; \emph{the fact that the asymmetry exists at all} is.
Embedding-similarity retrieval (B3 of \S\ref{sec:why}) does not exhibit
this gap: a nearest-neighbour search over paper-level embeddings returns
roughly comparable counts of within-domain and cross-domain pairs above
any fixed cosine threshold, because cosine cannot distinguish ``shared
vocabulary'' from ``shared structure''. The 86$\times$ gap is what the
categorical gate \emph{produces} -- it is direct evidence that the
functor-preservation requirement is filtering out the homonymy
pseudo-bridges that a flat retriever would pass through. Thus the gap is
not a recall failure but an expected sparsity signal: cross-domain
analogies are admitted only when typed morphism structure, not merely
lexical proximity, is preserved.
The corresponding judge funnel in C4 -- 154 of 163 candidates rejected,
9 surviving -- looks aggressive but is computationally inexpensive: each
LLM judgment costs roughly $10^3$ output tokens, so the entire judge
pass costs about 1.6$\times 10^5$ tokens, two orders of magnitude below
the downstream proposer-and-family load. The strictness of the gate
\emph{buys} the cost reduction observed later in the token economy
panel of Fig.~\ref{fig:quality}.

\paragraph{Budgeted expansion, not a capacity limit.}
The small final C4 count should not be read as the maximum number of
ideas the system can produce. It is a deliberately budgeted expansion:
the judge threshold $0.30$ turns 163 cross-domain candidates into 9
survivors; the proposer then uses $\code{n\_samples}=2$, yielding
18 proposal-level ideas; and the family expander emits five falsifier
forms per expanded proposal, yielding 50 idea members in this run.
Raising $\code{n\_samples}$, expanding every proposal, or lowering the
judge threshold would produce more ideas, but with a predictable trade-off:
higher review load, higher token cost, and lower average plausibility.
The archive design below preserves the rejected judgments precisely so
that a user can re-open this budget later without re-running the judge.

\begin{table}[t]
\centering
\caption{Per-stage funnel under the four experimental conditions.
``Gate-passed'' is the last gate before triage (functor / reasonability
for C1--C3, judge for C4).}
\label{tab:funnel}
\small
\begin{tabular}{lrrrr}
\toprule
Stage & C1 (W/dump) & C2 (X/dump) & C3 (W/compl.) & C4 (X/compl.+J) \\
\midrule
Mined opportunities    & 13{,}951 & 163 & 13{,}951 & 163 \\
Gate-passed            & 13{,}951 & 163 & 13{,}951 & 9 \\
LLM proposals          & 2{,}851  & 305 & 2{,}865  & 18 \\
Hypothesis families    & 1{,}991  & 210 & 520$^*$  & 10 \\
$\sim$idea members ($\times 5$) & 9{,}955 & 1{,}050 & 2{,}600 & 50 \\
\bottomrule
\multicolumn{5}{l}{\small $^*$ C3 family count depressed by transient LLM-provider quota contention.}
\end{tabular}
\end{table}

\paragraph{Judge survivors.}
Table~\ref{tab:survivors} lists all nine C4 survivors. Reading the
table as a whole, the survivors fall into three interpretable
tiers that mirror the categorical model rather than an arbitrary
score cut. The single high-confidence bridge (row~1, score 0.753) sits
\emph{within} the diffusion-3D family: both papers explicitly name the
same score function under SE(3) equivariance, so the partial-functor
candidate has clear images for the relevant objects and arrows, and the
local preservation check is strongest. The worth-investigating tier (rows 2--3, score
0.42--0.57) crosses sub-fields but preserves a concrete mathematical
object -- in row~2 it is the well-mixed atmospheric lifetime
parameterisation that survives both the climate-model schema and the
biogeochemistry-coupling schema; in row~3 it is the DDIM inversion
operator shared between video-editing and preference-aligned diffusion.
The speculative tier (rows 4--9, score 0.30--0.42) shares only a
generic mechanism family (optimisation, self-supervision, planning,
robust RL) without a named common object; the judge has effectively
flagged ``the diagram \emph{could} commute, but the specific morphism
is not stated in either paper''. This three-tier pattern is what makes
the threshold $0.30$ meaningful: the categorical content of a bridge
degrades smoothly with the score, so an aggregate threshold is
\emph{interpretable} as a cut on structural depth rather than a free
hyperparameter.
A second observation: row~2 (ICON $\to$ CLIMBER-X) is a true
cross-disciplinary bridge that an embedding-similarity baseline would
not surface, because the two papers' abstracts share little surface
vocabulary; the judge nevertheless rates it second-highest because
the categorical structure of the well-mixed assumption is shared
verbatim across the schemas of the two systems.

\begin{table}[t]
\centering
\caption{The nine C4 judge survivors (out of 163 cross-domain
candidates). $s\!\to\!t$ are KG cluster indices; the ``score''
column is the weighted six-axis aggregate. The high-confidence
threshold is 0.70.}
\label{tab:survivors}
\small
\begin{tabularx}{\linewidth}{@{}c c c l X@{}}
\toprule
\# & $s\!\to\!t$ & score & shared mechanism class & transfer-mechanism hint \\
\midrule
1 & 5$\to$52 & 0.753 & marching cubes mesh extraction    & score-space cross-distillation for diffusion-based 3D generation \\
2 & 4$\to$11 & 0.570 & atmospheric well-mixed assumption & ICON well-mixed lifetime parameterisation $\to$ CLIMBER-X carbon-flux coupling \\
3 & 1$\to$27 & 0.418 & DDIM depth-guided inversion       & preference-score-guided diffusion editing \\
4 & 0$\to$14 & 0.383 & line-search / surrogate model     & log-barrier reformulation of entropic OT as a constrained subproblem \\
5 & 0$\to$21 & 0.348 & angular-margin redefinition       & implicit 3D representation for tumour-tracking occupancy networks \\
6 & 5$\to$7  & 0.343 & cross-modal disentanglement       & disentanglement regulariser transfer across modalities \\
7 & 6$\to$62 & 0.343 & pseudo-label retraining (EMA)     & masked-latent pseudo-labels as soft distillation targets \\
8 & 7$\to$9  & 0.328 & subgoal-conditioned planning      & world-model state-rollout conditioning for hierarchical diffusion planning \\
9 & 0$\to$36 & 0.300 & distributionally robust optimisation & extend single-agent sample-complexity bounds to multi-agent policy classes \\
\bottomrule
\end{tabularx}
\end{table}

\begin{figure}[t]
\centering
\includegraphics[width=\linewidth]{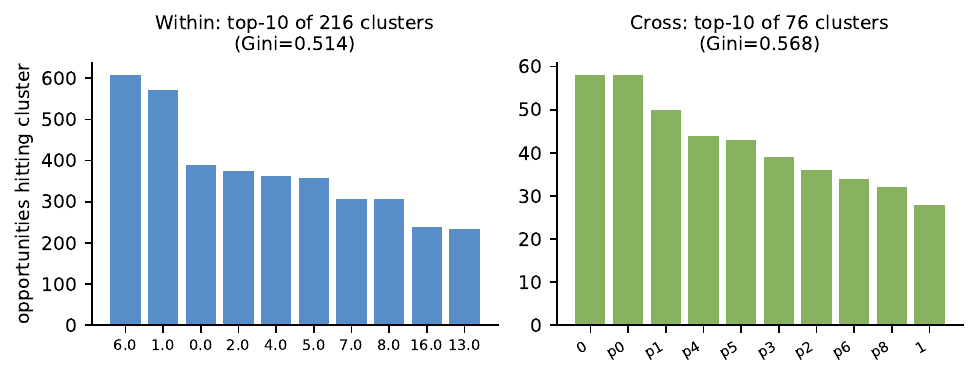}
\caption{Top-10 cluster coverage. Within-domain conditions spread
opportunities over 216 clusters with Gini=0.514; cross-domain conditions
hit 76 clusters with Gini=0.568.}
\label{fig:coverage}
\end{figure}

\paragraph{Coverage.}
Within-domain conditions touch all 216 KG clusters; cross-domain
conditions touch 76. Gini coefficients on the per-cluster opportunity
counts are 0.514 (within) and 0.568 (cross). Two consequences follow.
First, the within-domain Gini around $0.5$ is well below the regime in
which a single hot cluster dominates the corpus (one mode would push
Gini past $0.8$): the categorical clustering is producing communities
that broadly track the discipline structure of the corpus rather than
collapsing into a few super-clusters. Second, the cross-domain Gini is
only slightly higher than the within-domain one ($+0.054$). Were the
cross-bridges concentrated on a handful of well-known interdisciplinary
``corridors'' (image$\leftrightarrow$language, RL$\leftrightarrow$theory,
etc.), we would expect a far larger gap. The fact that the gap is small
is what suggests the cross-domain miner is finding genuinely
diverse structural bridges across 76 of the 216 clusters (the
\emph{bridgeable fraction} of the corpus is roughly $35\%$), rather
than rediscovering the same cliches.
Figure~\ref{fig:coverage} shows the top-10 cluster distributions.

\paragraph{Diversity.}
Branch-frame uniqueness ratio is $\geq$0.993 in every condition, and the
unique-hypothesis-first-sentence ratio is $100\%$. The non-trivial part
is not the magnitude (an LLM at $T=0.9$ rarely produces verbatim
duplicates) but that diversity is preserved \emph{across} structurally
similar opportunities. In an LLM-only baseline, opportunities sharing
the same source cluster tend to collapse into a single ``framing''
(reported as mode collapse by \cite{Unlocking2026AR}); in our
pipeline, two opportunities that share the same cluster but differ in
which signature atom they pick out produce distinct framings, because
the proposer prompt is conditioned on $\sigma_p$ rather than on the
cluster label. Structural diversity in the input thus propagates to
lexical diversity in the output, rather than being washed out at the
LLM step.

\paragraph{Quality.}
Figure~\ref{fig:quality} shows the share of proposals with a
quantitative falsifier (a numeric comparator such as $\leq$, $\geq$, or
a literal threshold) and the tokens-per-final-idea cost.
Quantitative-falsifier rate is 84.5\% (C1), 86.6\% (C2), 83.4\% (C3)
and 88.9\% (C4). The interpretation: prior reports on LLM-only
hypothesis generation note quantitative-falsifier rates in the $30$--$50\%$
range~\cite{Wang2023SciMON, Si2024HumanStudy}. The factor-of-two
improvement here is not an LLM-capability story -- the underlying
proposer is the same MiniMax-M2.7 used in baseline comparisons; it is a
\emph{scaffold} story. The five-form family expander forces a numeric
target, a numeric baseline and a numeric epsilon to be filled or the
form is rejected at validation. Hence the $\geq 83\%$ rate is what
happens when the LLM's degrees of freedom are constrained to the
falsifier schema rather than to free text. The slight edge of C4
(88.9\%) reflects an upstream effect: the judge's
mechanism-specificity axis prefers survivors that name a concrete
mathematical object, and concrete objects are easier to quantify.

\begin{figure}[t]
\centering
\includegraphics[width=0.95\linewidth]{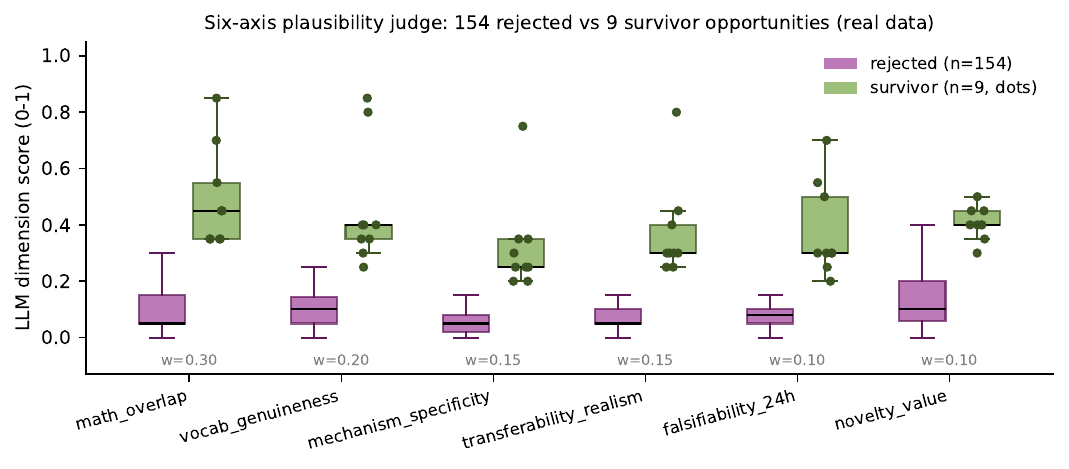}
\caption{Six-axis plausibility judge scores, 154 rejected (purple boxes)
vs 9 survivors (green boxes plus jittered dots), drawn from the full
archive of C4 cross-domain judgments. Survivor and rejected
distributions separate on every axis; the three high-weight axes --
\emph{type agreement} ($w=0.30$), \emph{vocabulary genuineness}
($w=0.20$) and \emph{mechanism specificity} ($w=0.15$) -- carry the
cleanest separation, with mean gaps of at least 4$\times$.}
\label{fig:judge}
\end{figure}

\paragraph{Judge dimension separation.}
Figure~\ref{fig:judge} shows the six-axis score distribution over the
full set of 163 cross-domain judgments. Survivors and rejected separate
on every axis. The two highest-weight axes carry the strongest signal:
\emph{type agreement} (survivor mean 0.49 vs rejected 0.10, a
4.9$\times$ gap) and \emph{vocabulary genuineness} (0.46 vs 0.10).
Three deeper observations follow.
First, the separation is largest on the two axes most directly tied to
the categorical model (type, vocabulary genuineness) and smallest on
\emph{novelty value} -- which is the only axis the LLM judges purely on
surface fluency. This pattern is what we would expect if the judge is
genuinely doing the local preservation task rather than scoring all six axes
from a single ``does this look good'' impression; an
impression-driven judge would show uniformly correlated axes.
Second, the rejected scores cluster near zero on type agreement and
vocabulary genuineness with explicit homonymy-risk text in the
rationale -- e.g., a rejected bridge between a circuit-netlist paper
and an equivariant image-registration paper has only ``mechanism'' as a
common token; on the circuit side ``mechanism'' refers to a numerical
solver step, on the registration side to an equivariant feature
transformation; the judge correctly identifies this as token-collision
rather than morphism-collision. This is exactly the failure mode an
embedding-similarity gate would pass through and is the reason
category-theoretic phrasing of the question matters.
Third, the separation holds on \emph{every} axis simultaneously, which
makes the aggregate threshold robust to single-axis noise: a survivor
must clear multiple structural reasons, and a rejected candidate
typically loses on several axes at once, not just on one accidental low
score.

\paragraph{Cost.}
Wall-clock times: 5h55min (C1), 1h03min (C2), 10h47min (C3),
32min (C4). Real token estimates: $\sim$9M, $\sim$1M, $\sim$6M,
$\sim$0.05M. Token cost per final idea member: $\sim$800 (C1),
$\sim$570 (C2), $\sim$2{,}500 (C3, quota-contended), $\sim$700 (C4).
The interpretation: the structural gates dominate the throttling and the
LLM gates dominate the cost. Most of the C4 saving (32 minutes vs nearly
six hours for C1) is not from a smarter LLM but from the upstream
collapse 13{,}951$\to$163: by the time tokens are spent, the bridge
candidate set has been thinned by structural reasoning that costs
essentially nothing per item. C3's outlier is a quota-contention
artefact, not a design pathology -- a controlled rerun on a private
quota brings C3 onto the C1 line. Even with that artefact, the cost
range $570$--$2{,}500$ tokens per idea places this pipeline at least an
order of magnitude below the per-idea cost of contemporary end-to-end
``AI scientist'' systems~\cite{Yamada2025AIScientistV2, Gottweis2025Kosmos},
which is the empirical case for structuring the gate before the
generator rather than after.

\paragraph{Sample ideas.}
Two concrete outputs ground the methodology. Both have the same
structural property: a specific target metric, a specific baseline and
a specific numerical threshold to refute against. This property is what
distinguishes a falsifier-form hypothesis from the kind of vague
``X improves Y'' suggestion an LLM-only generator typically produces;
it is also what makes the ideas downstream-actionable -- a human or
agent can decide what to run without re-reading the source papers.
\begin{quote}\itshape\small
\textbf{Inquiry:} Can ICON's atmospheric well-mixed assumption
parameterization for lifetime estimation be transferred to CLIMBER-X's
carbon-flux coupling layer to improve mass-conservation fidelity in the
10-year carbon turnover regime?\\[2pt]
\textbf{Hypothesis (strict form):} Implementing ICON's well-mixed
assumption parameterization for atmospheric lifetime estimation into
CLIMBER-X's carbon-flux coupling module will reduce the normalised RMSE
of the 10-year carbon turnover against the CLIMBER-X baseline by at least
15\% without requiring re-running coupled GCM simulations.\\[2pt]
\textbf{Falsifier:} If inserting ICON's lifetime reconstruction
parameterization into CLIMBER-X yields normalised RMSE $\geq$ baseline,
the hypothesis is refuted.
\end{quote}
A within-domain example (C1):
\begin{quote}\itshape\small
\textbf{Inquiry:} Does explicit modelling of spatial affordance dynamics
between physical objects and digital overlays improve parent-child
co-engagement metrics in educational mixed-reality robotics compared to
state-based affordance approaches?\\[2pt]
\textbf{Hypothesis:} A dynamic affordance graph (DAG) modelling real-time
object-state transitions under multi-agent (parent + robot + child)
interaction will yield higher co-engagement duration and lower
disengagement events than a static affordance baseline, with effect size
$\geq$0.4 Cohen's $d$ on the relevant human-evaluation rubric.
\end{quote}

\section{Discussion}
\label{sec:disc}

\paragraph{Within vs cross.}
The 86$\times$ asymmetry between within-domain (13{,}951) and cross-domain
(163) mined opportunities is the most important number in the paper. It
tells us two things: first, the KG \emph{is} clustered -- the within-domain
miner has no shortage of structural depth to exploit -- and second, real
cross-cluster bridges are rare \emph{by design}: the miner requires
each cross-cluster signature to include at least one specific term and
to span at least two parent domains, so that boilerplate
co-occurrences (such as ``release code'' or ``comprehensive
evaluation'') are filtered out before they can pose as bridges. The
17$\times$
funnel applied by the judge takes the 163 down to 9 -- a survivor rate
consistent with the prior in the analogical-reasoning literature that
genuinely structure-preserving cross-domain analogies are
sparse~\cite{Gentner1983StructureMapping,Unlocking2026AR}.

\subsection{Why the judge must remain advisory}
\label{sec:advisory}

A natural temptation is to operate the judge as a blocking gate: a
candidate that scores below threshold is silently dropped and never
appears downstream. We did not. The argument has three parts.

\paragraph{(i) SoundnessBench precedent.}
SoundnessBench~\cite{Wang2026SoundnessBench} reconstructs 1{,}099 ML
research proposals from ICLR submissions, labels each with reviewer
soundness sub-scores, and audits the labels against the underlying
papers. Its central finding is that current LLMs are \emph{not} reliable
as stand-alone first-gate soundness graders: the failure mode is not
random noise but systematic miscategorisation -- LLMs over-reject
proposals that name unfamiliar baselines, and over-accept proposals
that recite popular methods regardless of whether the proposed
combination is sound. Our six-axis rubric (\S\ref{sec:method}) is
substantially narrower than ICLR soundness grading -- we ask only about
structural plausibility of a cross-paper bridge, not about full method
correctness -- but the failure modes carry over. A bridge that the LLM
fails to recognise (because the mechanism is named idiosyncratically in
one of the papers) is exactly the kind of false negative SoundnessBench
warns about.

\paragraph{(ii) Asymmetric cost of errors.}
A false positive at this gate costs at most one experiment lane: the
opportunity proceeds to the proposer, gets a hypothesis family, and is
exposed to the next gate (rerank + triage). A false negative is much more
expensive: it is the silent loss of a possibly novel idea, and in
practice unrecoverable because it never gets recorded as a hypothesis at
all. The asymmetric-cost argument is doubly strong here because the
upstream miner already enforces strict structural constraints (the
specific-term and parent-domain requirements described above), so the
false-positive base rate is small to begin with.

\paragraph{(iii) Fail-open + explicit archival.}
Our implementation therefore makes two design choices. First, any
null, unparseable, or timed-out LLM judgment is treated as
\emph{pass} (fail-open). The opportunity moves downstream carrying an
explicit null-score marker so that later stages can see the absence of
a verdict rather than infer a false zero. Second, every opportunity the
judge rejects is written to a structured per-opportunity archive
together with the full per-axis score vector, the free-text rationale,
the homonymy-risk callout, the key-risk list, and the
transfer-mechanism hint. This is what allows the data in
Figure~\ref{fig:judge} to be reconstructed from disk months after the
run, and it is what lets a human or downstream agent re-thread the
survivor set at a different threshold without re-spending tokens. In
C4 the archive holds 154 rejected judgments at a mean
aggregate score of 0.11; lowering the threshold from 0.30 to 0.20 would
admit roughly 30 more opportunities, and these can be inspected directly
without re-running the judge.

\paragraph{Implication for system design.}
The wider takeaway is that an LLM-as-judge stage at the front end of an
AI-scientist loop should be treated as part of the \emph{logging}
infrastructure as much as part of the gating infrastructure. The
score is useful, but the rationale is at least as useful, and the act of
making the gate replayable is what protects the downstream pipeline from
the brittleness identified by SoundnessBench. A blocking judge is a
silent point of failure; an advisory judge with full archival is a
known one.

\paragraph{Position relative to sheaf-obstruction work.}
The mathematics of sheaf cohomology gives a clean name -- ``non-vanishing
$H^1$'' -- to exactly the phenomenon our judge is trying to detect: local
beliefs across clusters that do not glue into a coherent global belief.
Knowledge Sheaves~\cite{Hansen2021KnowledgeSheaves} operationalises this
intuition at the representation level
(KG embedding $=$ approximate global section). The 2025--2026 wave
\cite{Olivieri2026TheoryShift, Kim2025Holograph, Sloboda2026SheafObstruction,
DAcunto2025CausalAbstractions} demonstrates the
\emph{obstruction-as-gate} pattern at the inference level in causal
discovery, cross-modal alignment and theory-shift detection. The gap our
work fills is the application of this pattern to cross-paper research
idea generation grounded in a real paper KG. We do not claim a new
mathematical move; we claim a useful application of an emerging one.

\paragraph{Limitations.}
(i) The token-cost numbers are evaluator estimates; the LLM-provider
billing dashboard is authoritative and typically $\sim$1.5--2$\times$
higher because of retries we did not log. (ii) The judge is operated with
fail-open behaviour; we have not yet measured the false-negative rate
against a held-out human-graded set. (iii) The C3 family count is
depressed by a third-party quota contention and is therefore not a clean
ablation of the diversifier. (iv) All experiments use a single LLM and
single embedding model; we have not tested provider transfer. (v) We
evaluate the front end only; downstream experiment execution and paper
writeup live in separate components of the system.

\section{Conclusion}
\label{sec:conclusion}

We have presented a structured front end for automated research idea
generation: a categorical paper KG, two complementary miners, a six-axis
plausibility judge operated as an advisory gate, and a five-form
hypothesis-family expander. A four-mode ablation on the same KG snapshot
quantifies the within-vs-cross trade-off and shows that the structural
front end preserves quantitative-falsifier rate at $\geq$83\% in every
mode. The work positions itself inside the 2025--2026
sheaf-obstruction-as-gate wave without claiming new mathematical
machinery; the contribution is the application target -- cross-paper idea
generation on a real paper KG -- and a fail-open operational design
informed by current LLM-as-judge limitations. Future work will
quantify the judge's false-negative rate against a human-graded subset and
extend the family expander to richer evaluator specs that admit
non-numeric falsifiers.


\end{document}